\documentclass[english]{article}
\usepackage[T1]{fontenc}
\usepackage[latin9]{inputenc}
\usepackage{amstext}
\usepackage{amssymb}
\usepackage{graphicx}
\usepackage{esint}

\makeatletter

\providecommand{\tabularnewline}{\\}

\usepackage{spconf}
\name{Kevin R. Moon$^{1}$ \qquad Jimmy J. Li$^{1}$  \qquad V\'{e}ronique Delouille$^2$ \qquad Fraser Watson$^{3}$  ~~~~~Alfred O. Hero III$^1$\thanks{This work was partially supported by NSF grant CCF-1217880 and a NSF Graduate Research Fellowship to the first author under Grant No. F031543.}}

\address{$^{1}$ University of Michigan, Dept. of EECS, 1301 Beal Avenue, Ann Arbor, MI 48109, USA\\
    $^{2}$ Royal Observatory of Belgium, SIDC, Avenue Circulaire 3, 1180 Uccle, Belgium\\
$^3$National Solar Observatory, 950 N. Cherry Ave, Tucson, AZ 85719, USA}
\makeatother

\usepackage{babel}
\begin{document}

\title{Image patch analysis and clustering of sunspots: a dimensionality
reduction approach}
\maketitle
\begin{abstract}
\ninept

Sunspots, as seen in white light or continuum images, are associated
with regions of high magnetic activity on the Sun, visible on magnetogram
images. Their complexity is correlated with explosive solar activity
and so classifying these active regions is useful for predicting future
solar activity. Current classification of sunspot groups is visually
based and suffers from bias. Supervised learning methods can reduce
human bias but fail to optimally capitalize on the information present
in sunspot images. This paper uses two image modalities (continuum
and magnetogram) to characterize the spatial and modal interactions
of sunspot and magnetic active region images and presents a new approach
to cluster the images. Specifically, in the framework of image patch
analysis, we estimate the number of intrinsic parameters required
to describe the spatial and modal dependencies, the correlation between
the two modalities and the corresponding spatial patterns, and examine
the phenomena at different scales within the images. To do this, we
use linear and nonlinear intrinsic dimension estimators, canonical
correlation analysis, and multiresolution analysis of intrinsic dimension.
\end{abstract}
\ninept

\section{Introduction}

Sunspots are associated with active regions, which are areas of locally
increased magnetic flux on the Sun. The morphology of sunspot groups
and associated active regions is correlated with the incidence of
solar flares~\cite{Bornmann1994flares}. The current practice for
identifying and classifying sunspot groups is based on the Mount Wilson
classification scheme, which categorizes them by eye based on morphological
criteria present in continuum and magnetogram images. Such visual
classification introduces bias stemming from the artificial and subjective
nature of the discrete categorization. It also makes the study of
the sunspot group's dynamic behaviour impractical. 

Recent works~\cite{2008SoPh..248..277C,2013Stenning} have attempted
to reproduce the Mount Wilson classification through automated procedures
while~\cite{Ireland2008mra} has employed multiresolution analysis
to differentiate the various types of active regions. While these
approaches reduce the human bias, they do not use the information
present in sunspot images in an optimal way. This paper presents for
the first time a spatial correlation and intrinsic dimension analysis
of sunspot images and a new approach to cluster the images. We use
two image modalities (continuum and magnetogram) to characterize their
spatial and modal interactions for improved sunspot classification.
To do this, we first address three questions for umbral and penumbral
regions of the sunspot. 1) How many intrinsic parameters or degrees
of freedom are required to describe the spatial and modal dependencies?
2) What correlation exists between the two modalities and what spatial
patterns produce that correlation? 3) What phenomena exist at different
scales within the images? We use this information to cluster the images
by clustering dictionaries learned from each image.

The paper is organized as follows. In Sec.~\ref{sec:data}, a description
of the dataset is provided. In Sec.~\ref{sec:dimension}, the intrinsic
dimension of the joint image is estimated using both nonlinear and
linear methods. We also perform a multiresolution analysis (MRA) of
intrinsic dimension. We then identify complex spatial and modal interactions
at different scales that are not visible to the naked eye by using
canonical correlation analysis (CCA) in Sec.~\ref{sec:correlation}.
Section~\ref{sec:clustering} then presents our clustering approaches
and results.

MRA has been used for many image applications including denoising
and reconstruction~\cite{willett2003platelets}, segmentation~\cite{unser1989segmentation},
and representation~\cite{jansen2005multiscale}. Many of these methods
use a basis transformation and then a linear decomposition of the
transformed data. Our case differs in that the sunspot images are
vector valued (two modalities), possibly non-linear, and nonstationary
so standard linear MRA may not be sufficient to capture the interactions
between the two modalities.

\section{Data}

\label{sec:data}The data used in this study are taken from the MDI
instrument~\cite{1995MDI} on board the SOHO Spacecraft. The active
regions of the Sun are observed using level 1.8 continuum (cont) and
level 1.8 magnetogram (mag) images. Active regions are selected within
$30$ degrees of the solar meridian to avoid strong projection effects.
Expertly generated masks marking the location of the umbra and penumbra
of the sunspots are available for each set of images~\cite{2011masks}.
Information about the sunspot groups such as Mount Wilson class labels,
Zurich class labels, and sunspot group longitudinal extent comes from
the Solar Region Summary reports compiled by the Space Weather Prediction
Center of NOAA~http://www.swpc.noaa.gov/ftpdir/forecasts/SRS/.

\begin{figure}
\centering

\includegraphics[width=0.25\textwidth]{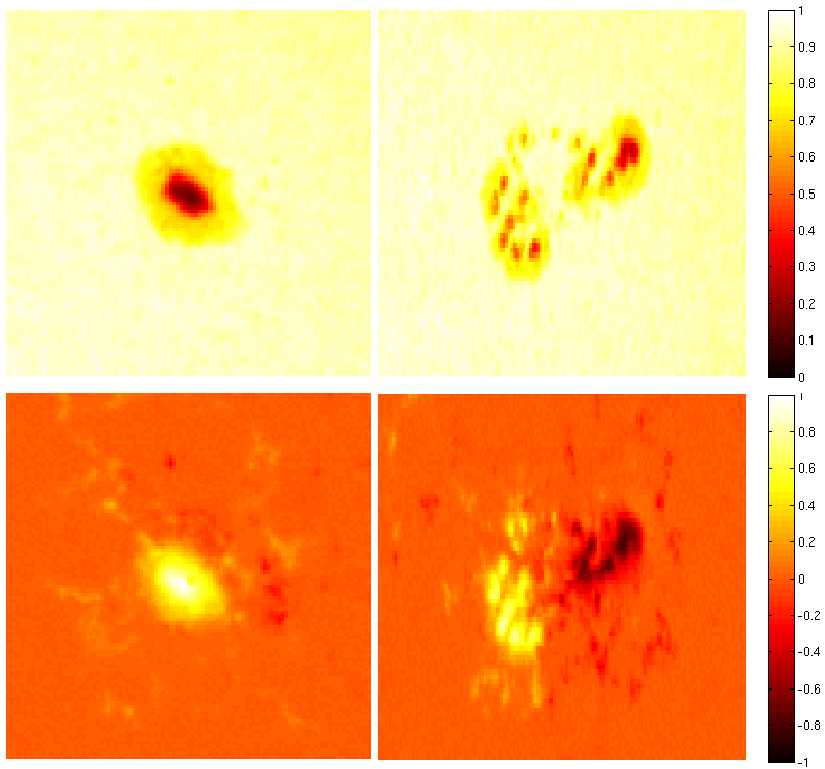}

\caption{Cont (top) and mag (bottom) images. Left to right: single spot, multiple
spots. \label{fig:images}}
\end{figure}

\begin{figure}
\centering

\includegraphics[width=0.2\textwidth]{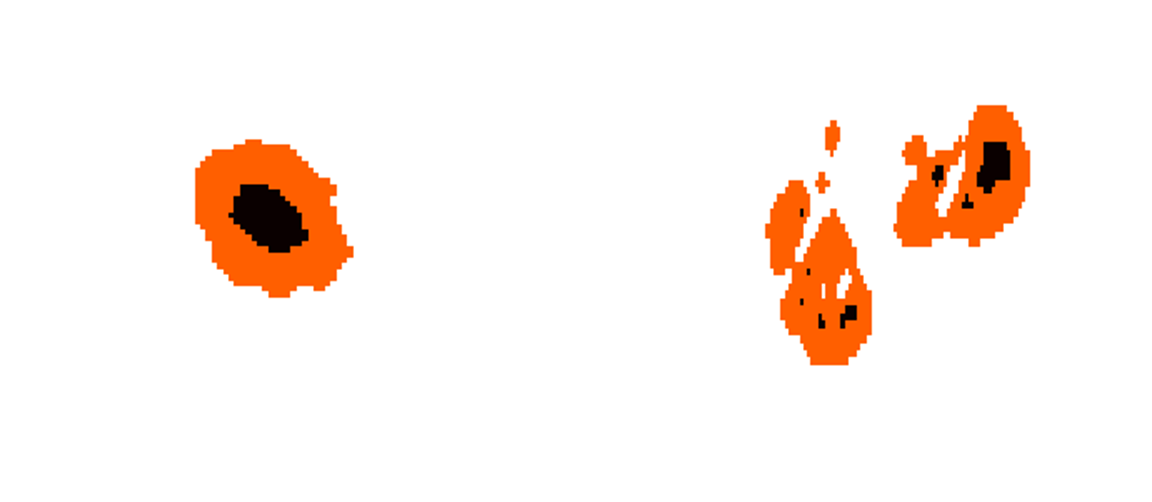}

\caption{Masks for the single spot and multiple spot images extracted by solar
sunspot experts. Interior = umbra, exterior = penumbra.\label{fig:masks}}
\end{figure}

Similarly to \cite{malik2001contour,mairal2008discriminative,fergus2004visual},
we use image patch features to account for spatial dependencies using
square patches of pixels. Thus if an image has $n$ pixels and we
use a $3\times3$ patch, the corresponding cont data matrix $X$ is
$9\times n$ where the $i$th column contains the pixels in the patch
centered at the $i$th pixel. The mag data matrix $Y$ is formed in
the same way and the full data matrix is $Z=\left(\begin{array}{cc}
X^{T} & Y^{T}\end{array}\right)^{T}$. 

While we have applied our analysis to a large corpus of sunspot images,
two specific images are used to illustrate our results: a region with
a single sunspot and a region with intense magnetic activity and multiple
sunspots. The images and masks are given in Figs.~\ref{fig:images}
and~\ref{fig:masks}, respectively.

\section{Intrinsic Dimension Estimation}

\label{sec:dimension}To determine the number of intrinsic parameters
or degrees of freedom required to describe the spatial and modal dependencies,
we estimate the local intrinsic dimension of the joint $3\times3$
patches, which lie in an extrinsic Euclidean space of 18 dimensions.
We investigate two different methods for estimation of intrinsic dimension:
the first is appropriate to linear subspaces while the second is appropriate
to any (linear or non-linear) smooth subspace. The linear method we
use is principal component analysis (PCA). PCA finds a set of linearly
uncorrelated vectors (principal components) that can be used to represent
the data. The principal components are the eigenvectors of the covariance
matrix $\Sigma=\left(\begin{array}{cc}
\Sigma_{\mathbf{xx}} & \Sigma_{\mathbf{xy}}\\
\Sigma_{\mathbf{yx}} & \Sigma_{\mathbf{yy}}
\end{array}\right),$ where $\mathbf{x}$ and $\mathbf{y}$ are random vectors, $\mathbf{x}$
is a patch from the cont image, and $\mathbf{y}$ is the corresponding
patch from the mag image. The eigenvalues indicate the amount of variance
accounted for by the corresponding principal component. A linear estimate
of intrinsic dimension is the number of principal components that
are required to explain a certain percentage of the variance. To account
for differences between the umbra, penumbra, and background, PCA is
performed separately on those areas using the masks provided in Fig.~\ref{fig:masks}.

The nonlinear method we use is a $k$-NN graph approach with neighborhood
smoothing~\cite{carter2010local} which is as follows. For a set
of independently identically distributed random vectors $\mathbf{Z}_{n}=\{z_{1},\dots,z_{n}\}$
with values in a compact subset of $\mathbb{R}^{d}$, the $k$-nearest
neighbors of $z_{i}$ in $\mathbf{Z}_{n}$ are the $k$ points in
$\mathbf{Z}_{n}\backslash\{z_{i}\}$ closest to $z_{i}$ as measured
by the Euclidean distance $||\cdot||$. The $k$-NN graph is then
formed by assigning edges between a point in $\mathbf{Z}_{n}$ and
its $k$-nearest neighbors and has total edge length defined as 
\[
L_{\gamma,k}(\mathbf{Z}_{n})=\sum_{i=1}^{n}\sum_{z\in\mathcal{N}_{k,i}}||z-z_{i}||^{\gamma},
\]
 where $\gamma>0$ is a power weighting constant and $\mathcal{N}_{k,i}$
is the set of $k$ nearest neighbors of $z_{i}.$ The asymptotics
of $L_{\gamma,k}(\mathbf{Z}_{n})$ are given in the following theorem~\cite{costa2006determining}: 

\newtheorem{Convergence}{Theorem}
\begin{Convergence}
Let $(\mathcal{M},g)$ be a compact smooth Riemann $m$-dimensional manifold. Suppose $z_1,\dots,z_n$ are i.i.d. random elements of $\mathcal{M}$ with bounded density $f$ relative to $\mu_g$. Assume that $m\geq 2$, $1\leq\gamma<m$ and define $\alpha=(m-\gamma)/m$. Then

\[
\lim_{n\rightarrow\infty}\frac{L_{\gamma,k}(\mathbf{Z}_{n})}{n^{\alpha}}=\beta_{m,L_{\gamma}}\int_{\mathcal{M}}f^{\alpha}(z)\mu_{g}(dz)\text{ }a.s.,
\]
 where $\beta_{m,L_\gamma}$ is a constant independent of $f$ and $\mathcal{M}$.
\end{Convergence}

Theorem 1 says that the total edge length of the $k$-NN graph increases
in $n$ at a sublinear rate $n^{\alpha}$ with $\alpha<1$, where
$\alpha$ is related to the intrinsic dimension $m$ of the manifold
$\mathcal{M}.$ The sublinear slope is closely related to the R\'{e}nyi
entropy $H_{\alpha}(f)=(1-\alpha)\ln\int_{\mathcal{M}}f^{\alpha}(z)\mu_{g}(dz)$
of the density $f$ on the manifold. Then for large $n$:
\[
L_{\gamma,k}(\mathbf{Z}_{n})=n^{\alpha(m)}c+\epsilon_{n},
\]
 where $c$ is a constant with respect to $\alpha(m)$ that depends
on the Renyi entropy of the distribution of the manifold and $\epsilon_{n}$
is an error term that decreases to zero a.s. as $n\rightarrow\infty$
\cite{costa2006determining}. Using this expression, a global intrinsic
dimension estimate $\hat{m}$ is found using non-linear least squares
over different values of $n$ \cite{carter2010local}.

To find a local estimate of dimension at a point $z_{i}$, the algorithm
is run over a smaller neighborhood about $z_{i}.$ However, this can
result in highly variable estimates of dimension for nearby points.
This variance can be reduced by smoothing the intrinsic dimension
estimate by majority voting in a neighborhood of $z_{i}.$ Specifically,
\[
\hat{m}(z_{i})=\hat{m}(i)=\arg\max_{l}\sum_{z_{j}\in\mathcal{N}_{i}}1(\hat{m}(z_{j})=l),
\]
 where $1(\cdot)$ is the indicator function and $\mathcal{N}_{i}$
is the neighborhood of $z_{i}$ \cite{carter2010local}. We use $|\mathcal{N}_{i}|=6$. 

\begin{figure}
\centering

\begin{centering}
\includegraphics[width=0.25\textwidth]{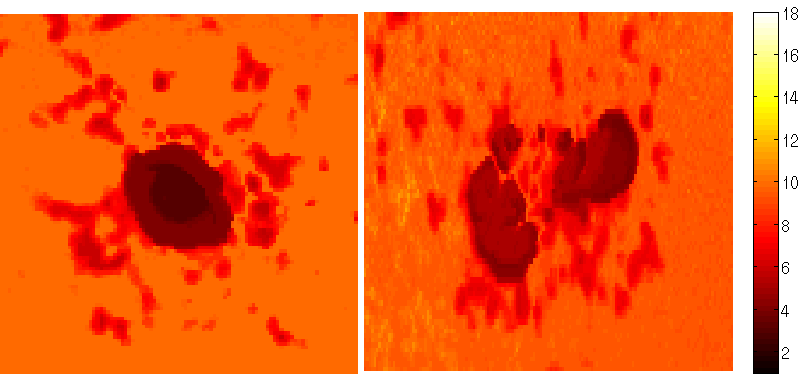}
\par\end{centering}

\caption{$k$-NN estimate of local intrinsic dimension ($\hat{m}(i)$, where
$i$ indexes over image pixels) of the single sunspot (left), and
multiple sunspot (right) images.  The standard deviation across iterations
for each pixel is generally less than 1. \label{fig:dimension}}
\end{figure}

To account for variance due to random paths, we run the algorithm
20 times per image. Figure \ref{fig:dimension} gives the mean of
the estimated local dimension for each image using a $3\times3$ patch.
Most of the background of the single sunspot image has estimated dimension
varying between $\hat{m}=9$ or 10, which is consistent with estimates
obtained throughout pure background images (not shown). However, there
are regions with magnetic activity outside of the main sunspot (magnetic
fragments) that have lower estimated dimension $\hat{m}\approx5,$
or $6$. The sunspot also has lower dimension ($\hat{m}\approx3-5$)
than the background. This is expected since the background has less
structure compared to the sunspots and magnetic fragments. Similar
results are obtained for the multi-spot image.

Table~\ref{tab:dimension} gives the estimated intrinsic dimension
for 20 images (10 with a single spot, 10 with multiple spots) extracted
from the corpus using both PCA and the $k$-NN method. For the $k$-NN
method, the algorithm is run 20 times per image and the mean is reported
for each region. For PCA, the intrinsic dimension based on a 97\%
threshold averaged across images is reported. With the exception of
the umbra for single sunspots, the 97\% PCA result is within one standard
deviation of the $k$-NN mean. Thus depending on the precision required,
linear methods may be sufficient to represent the spatial and modal
dependencies within most of the images. PCA and $k$-NN estimates
are in closer agreement for multiple sunspot images than for single
sunspots. Since the multiple sunspot images often have more magnetic
fragments in the background than single sunspots, this suggests that
linear methods may perform better at representing these regions compared
to pure background. 

\begin{table}
\centering

\begin{tabular}{|l|c|c|c|}
\hline 
 & \multicolumn{1}{c|}{Background} & \multicolumn{1}{c|}{Penumbra} & \multicolumn{1}{c|}{Umbra}\tabularnewline
\hline 
\hline 
Single Spot $k$-NN & 8.9 & 4.5 & 3.4\tabularnewline
\hline 
Single Spot PCA & 10.1 & 4.3 & 6.3\tabularnewline
\hline 
Multiple Spots $k$-NN & 8.6 & 4.8 & 4.0\tabularnewline
\hline 
Multiple Spots PCA & 8.9 & 4.8 & 3.4\tabularnewline
\hline 
\end{tabular}\caption{Estimated intrinsic dimension for multiple images with single sunspots
and multiple sunspots using $k$-NN or PCA. PCA values correspond
to a 97\% threshold. \label{tab:dimension}}

\end{table}

\begin{figure}
\centering

\includegraphics[width=0.35\textwidth]{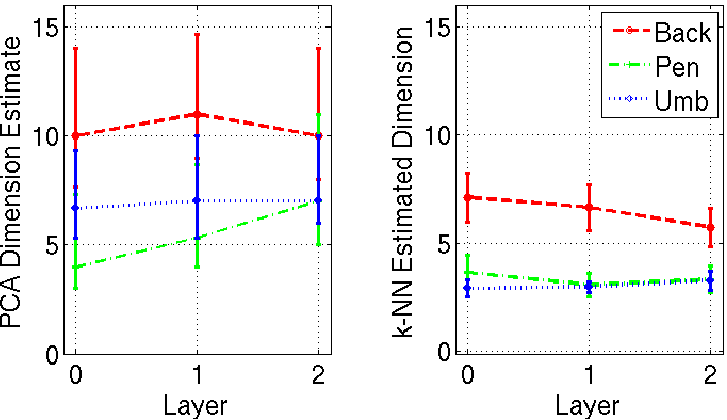}

\caption{Estimated intrinsic dimension using PCA (left) and the average $\hat{m}$
using the $k$-NN method (right) within each region as a function
of scale for three single sunspot images. The error bars correspond
to a single standard deviation for the $k$-NN method and to the 95\%
and 99\% thresholds for PCA (plot corresponds to 97\%). \label{fig:dimlayer}}
\end{figure}

To explore the existence of different phenomena at different scales,
we perform MRA on intrinsic dimension since the intrinsic dimension
estimates indicate the areas where the two modalities are most correlated.
Each scale (layer) is produced using a Haar wavelet decomposition
and reconstruction. We estimate the intrinsic dimension at each layer
for single sunspots using the two methods discussed in Sec.~\ref{sec:dimension}.
Both methods are used with $3\times3$ patches. At the 0th and 1st
layers, we average the results from three similar images. At the 2nd
layer, we analyze the combined data to ensure enough samples within
each region. The results using both PCA and the $k$-NN method are
given in Fig.~\ref{fig:dimlayer}. 

The results for the initial resolution are consistent with Table~\ref{tab:dimension}.
As scale increases, $\hat{m}$ decreases within the background while
the PCA estimate increases initially. Using the Jonckheere-Terpstra
trend (Jtrend) test~\cite{wolfe1973nonparametric} shows that both
relationships are significant. Since noise generally has a higher
dimension, this suggests that increasing the scale in the background
effectively denoises the data for the $k$-NN estimate. Within the
penumbra, $\hat{m}$ decreases initially and then increases while
the PCA estimate increases consistently. Within the umbra, both $\hat{m}$
and the PCA estimate increase gradually. The Jtrend test shows that
all of these relationships are significant except for the umbra PCA
estimate. This is similar to~\cite{georgiadis2013texture} where
the entropy of certain image textures is found to be generally increasing
but nonmonotonically with scale. 

\begin{figure}
\centering

\includegraphics[width=0.4\textwidth]{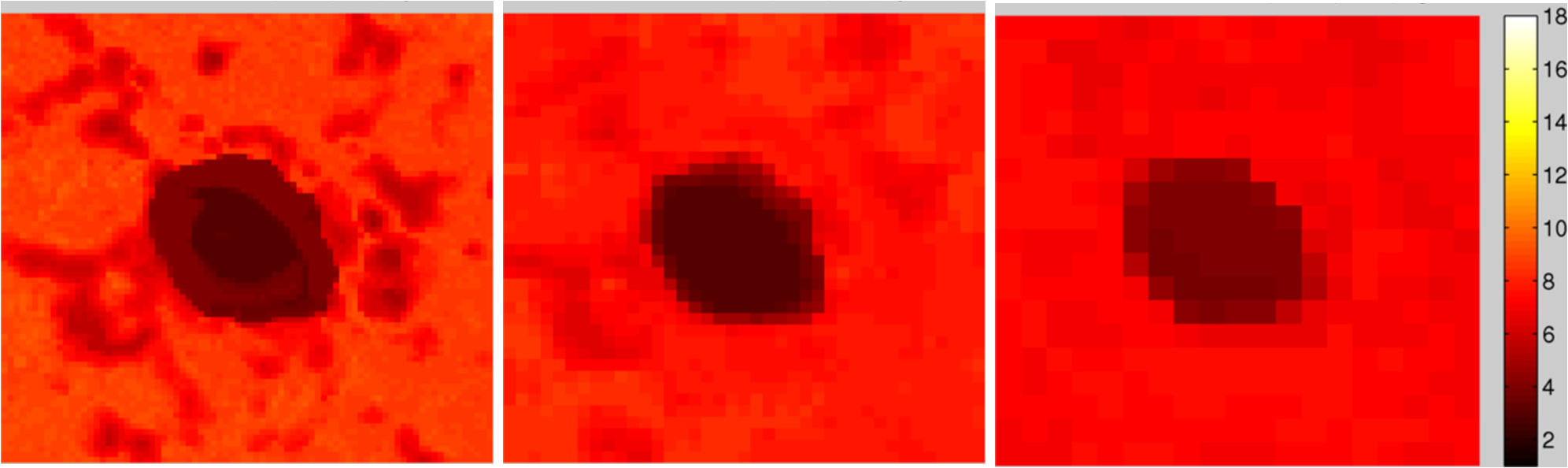}

\caption{Local estimated intrinsic dimension ($\hat{m}(i)$) of the single
sunspot image at different scales. L to R: 0th, 1st, and 2nd layers.\label{fig:dimMRA}}
\end{figure}

Figure~\ref{fig:dimMRA} shows $\hat{m}$ of the single sunspot image
at the different scales. The background in the 1st layer appears to
be a denoised version of the estimate at the original resolution.
Some of the magnetic fragments are preserved and the remaining background
is more uniform. Other trends in the images are consistent with Fig.~\ref{fig:dimlayer}.
Similar results are obtained for images with multiple sunspots.

\section{Correlation of Cont and Mag Images}

\label{sec:correlation}As the results in the previous section indicate
that linear methods may be sufficient to represent the spatial and
modal dependencies within a sunspot, we analyze the linear correlation
over patches. We do this by using canonical correlation analysis (CCA).
We perform this analysis within the background, the umbra, and the
penumbra of each image while using different patch sizes.

CCA finds vectors $a_{i}$ and $b_{i}$ such that the correlation
$\rho_{i}=\text{corr}(a_{i}^{T}\mathbf{x},b_{i}^{T}\mathbf{y})$ is
maximized and the pair of random variables $u_{i}=a_{i}^{T}\mathbf{x}$
and $v_{i}=b_{i}^{T}\mathbf{y}$ are uncorrelated with all other pairs
$u_{j}$ and $v_{j}$, $j\neq i$. The variables $u_{i}$ and $v_{i}$
are called the $i$th pair of canonical variables. The solution $a_{i}$
is the $i$th eigenvector of the matrix $\Sigma_{\mathbf{xx}}^{-1}\Sigma_{\mathbf{xy}}\Sigma_{\mathbf{yy}}^{-1}\Sigma_{\mathbf{yx}}$.
The vector $b_{i}$ is found similarly~\cite{hardle2007applied}.

\begin{figure}
\centering

\includegraphics[width=0.35\textwidth]{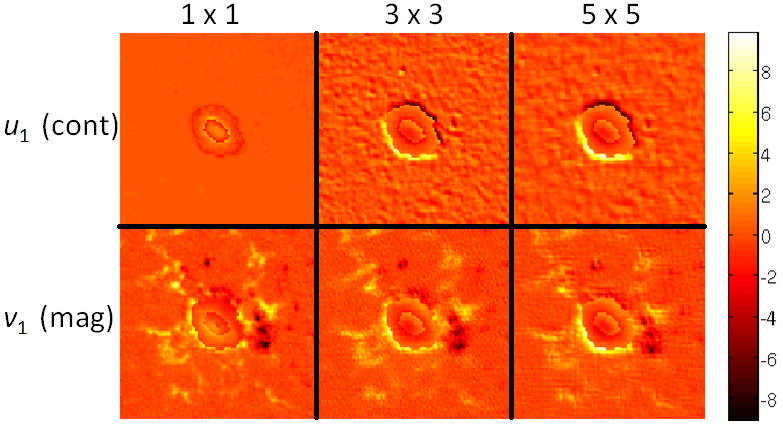}

\caption{Canonical variable images using different patch sizes of the single
sunspot image. Left to right: $1\times1$, $3\times3$, and $5\times5$
patch. $u_{1}$ (top) and $v_{1}$ (bottom). \label{fig:sunspot_canon}}
\end{figure}

The variables $u_{1}$ and $v_{1}$ for the single sunspot image using
different patch sizes are given in Fig.~\ref{fig:sunspot_canon}.
$u_{1}$ and $v_{1}$ are calculated separately for the background,
umbra, and penumbra and the first canonical correlation is approximately
$0.25$, $0.95$, and $0.9$ respectively using a $3\times3$ patch.
The areas with highest correlation (in magnitude) are primarily around
the edges of the penumbra and umbra as well as the magnetic fragments.
Some of the magnetic fragments are highly positively correlated while
others are highly negatively correlated. This correlation suggests
that classification algorithms should process both modalities together
for optimal performance.

As patch size is increased, the contrast in the images generally increases
at the expense of blurred edges. Thus multiple patch sizes may be
used to identify the regions with greatest correlation. 

\section{Clustering of sunspot images}

\label{sec:clustering}We applied the image patch analysis discussed
in Secs.~2-4 to unsupervised classification of sunspot images over
the corpus of sunspot images. First, we extract a 320 square pixel
region centered on each sunspot group. This results in 509 sunspot
group images taken from about 400 pairs of images. We then form a
data matrix for each sunspot group image using either pixel patches
as described previously or the canonical variables $u_{i}$ and $v_{i}.$
Next we use dictionary learning on each individual data matrix. Our
intrinsic dimension analysis of the corpus in Sec.~\ref{sec:dimension}
implies that the umbra, penumbra and magnetic fragments of the images
can be well approximated linearly (via PCA) by 7 dimensions or less.
Hence we use linear dictionary learning models and restrict the size
of the dictionaries to be less than or equal to 7. We then treat each
learned dictionary as a single vector and use spectral clustering
methods on these vectors to classify the images into distinct groups.

The clustering algorithm we use is the EAC-DC method in~\cite{galluccio2013clustering}
which scales well for clustering in high dimensions. EAC-DC clusters
the data by using a metric based on the hitting time of two Minimal
Spanning Trees (MST) grown sequentially from a pair of points. Consensus
spectral clustering is then applied to an ensemble of the resulting
dual rooted MSTs. This method was found to be robust and competitive
with other clustering algorithms~\cite{galluccio2013clustering}. 

To aid in interpreting our clustering results, we compare them to
the Mount Wilson labels with five classes: beta (1), alpha (2), beta-gamma
(3), beta-gamma-delta (4), and beta-delta (5). The normalized mutual
information (NMI) and adjusted Rand index (ARI) of the Mount Wilson
labels and our results using PCA to learn the dictionary from image
patches are $0.11$ and $0.03$, respectively. This low correspondence
is expected since our clustering approach is based on local image
patch features while the Mount Wilson labeling scheme focuses on global
features e.g. the polarities present in the group and whether they
can be separated spatially with a line. 

\begin{figure}

\includegraphics[width=0.5\textwidth]{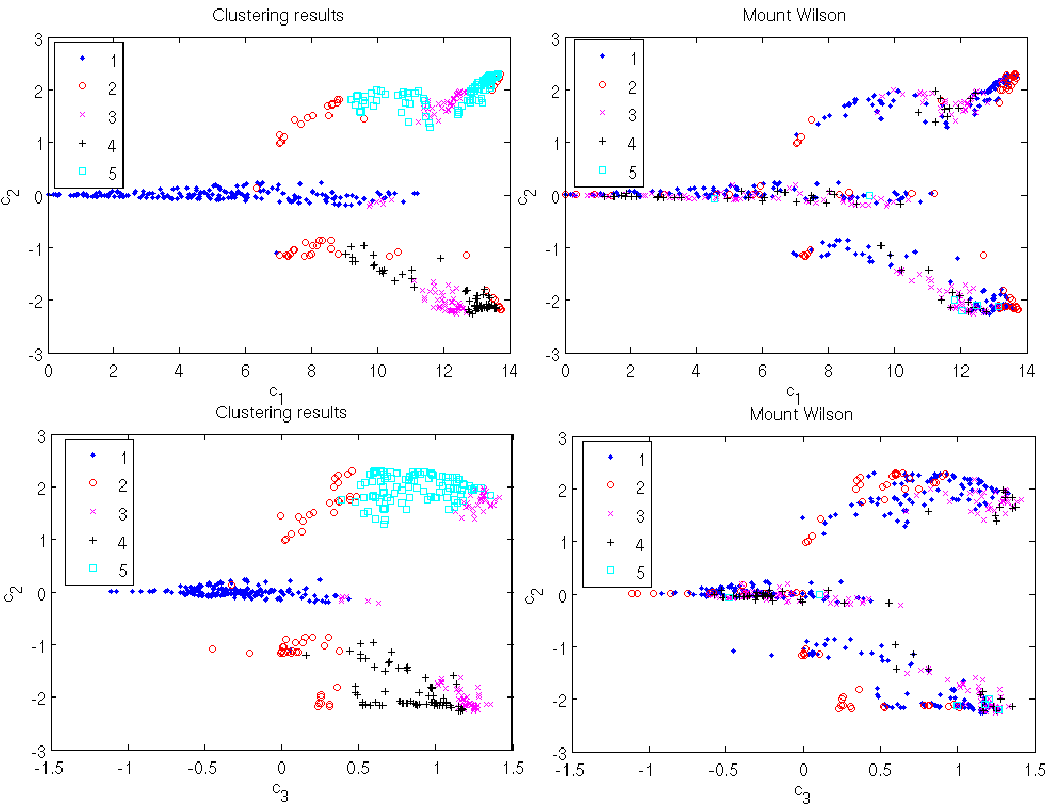}\caption{Plots of $c_{1}$ vs. $c_{2}$ (top) and $c_{3}$ vs. $c_{2}$ (bottom)
where $c_{i}$ is the projection of the similarity matrix onto the
$i$th eigenvector of the normalized Laplacian. Points are labeled
according to our clusters using the EAC-DC algorithm (left) and the
Mount Wilson labels (right). This shows that the intrinsic spatial
features are quite different from those used in the Mount Wilson classifications.\label{fig:clusters}}

\end{figure}

To visualize the images in low dimension, we projected the similarity
matrix created by the dual rooted MSTs onto the eigenvectors of the
normalized Laplacian of the similarity matrix, i.e. multidimensional
scaling~\cite{borg2005modern}. Figure~\ref{fig:clusters} gives
a scatter plot of $c_{1}$ vs. $c_{2}$ (top) and $c_{3}$ vs. $c_{2}$
(bottom) where $c_{i}$ is the projection onto the $i$th eigenvector.
The points are labeled according to the clusters (left) and the Mount
Wilson labels (right). The plots show that three clear groups of points
are clearly visible and linearly separable. The plots on the left
show that clusters 1, 4, and 5 are connected tightly while clusters
2 and 3 appear to be disconnected. In contrast, the Mount Wilson labels
are mixed throughout the three point clouds. However, there are still
some patterns present. For example, there are small groups of alpha
images (labeled 2) located on the left and right ends of the top and
bottom point clouds in the top right plot suggesting that our clustering
method finds distinct features of the images.

There is also some correlation with our results and the longitudinal
extent of the sunspot group: cluster 2 includes only smaller groups
with longitudinal extent less than 5 degrees while cluster 3 includes
only medium to large groups (extent greater than 6 degrees). Comparing
our results to the Zurich class labels, which depend more on longitudinal
extent, gives NMI and ARI of $0.16$ and $0.05$ respectively which
is slightly higher. This further demonstrates the value of our clustering
approach as it has some physical interpretability.

We also clustered the images by learning the dictionary on the canonical
variables using PCA and by learning the dictionary on the image patches
using the method in~\cite{ramirez2012mdl}. The results from these
approaches correspond even less with the Mount Wilson labels (NMI$<0.08$,
ARI$<0.03$) although all three methods resulted in little correspondence
(NMI, ARI$\approx0$) with each other. This is expected since CCA
only focuses on those regions where the two modalities are highly
correlated while the method in~\cite{ramirez2012mdl} can result
in dictionaries with different sizes. Viewing plots of the projections
$c_{i}$ as in Fig.~\ref{fig:clusters} also shows clearly separable
clusters indicating that combining these methods may result in improved
clustering.

\section{Conclusion}

We found the intrinsic dimension of the joint continuum and magnetogram
patches to be lower within the sunspot and magnetic fragments than
in the background suggesting stronger spatial and modal correlations.
CCA indicates that the areas that are most coupled are the magnetic
fragments and the transition regions between background, penumbra,
and umbra. Further work is required to evaluate the magnetic fragments
systematically.

The projections of the similarity matrix onto the eigenvectors of
the normalized Laplacian show that the mapping of the image dictionaries
using the dual rooted MSTs results in clearly separable regions which
can be clustered. While the NMI and ARI of the clustering results
and the Mount Wilson classes is low, some patterns are present. Also,
the clustering of image patch dictionaries using PCA is related somewhat
with the longitudinal extent of the sunspot groups suggesting some
physical interpretability. Future work includes clustering using combined
image patch and CCA data as well as global and long range spatial
features such as those used for the Mount Wilson scheme.

\bibliographystyle{IEEEbib}
\bibliography{sunspot_icassp}

\end{document}